\documentclass[10pt,twocolumn,letterpaper]{article}

\usepackage{cvpr}              
\usepackage{graphicx}
\usepackage{multirow}
\usepackage{amsfonts,amssymb}
\usepackage{amsmath}
\usepackage[mathscr]{eucal}
\usepackage{makecell}
\usepackage{marvosym}
\usepackage[dvipsnames]{xcolor}

\definecolor{cvprblue}{rgb}{0.21,0.49,0.74}
\usepackage[pagebackref,breaklinks,colorlinks,citecolor=cvprblue]{hyperref}

\title{Triple Disentangled Representation Learning for Multimodal Affective Analysis}

\author{Ying Zhou $^{\star}$, Xuefeng Liang \textsuperscript{\Letter} $^{\star}$, Han Chen $^{\star}$, Yin Zhao $^{\ast}$, Xin Chen $^{\star}$, Lida Yu $\dagger$ \\
$^{\star}$  School of Artificial Intelligence, Xidian University, China\\
	$^{\ast}$ Alibaba Group, Beijing, China\\
	$\dagger$ Beijing Normal University, Beijing, China\\
\textsuperscript{\Letter} xliang@xidian.edu.cn}

\begin{document}
\maketitle
\begin{abstract}
	Multimodal learning has exhibited a significant advantage in affective analysis tasks owing to the comprehensive information of various modalities, particularly the complementary information. Thus, many emerging studies focus on disentangling the modality-invariant and modality-specific representations from input data and then fusing them for prediction. However, our study shows that modality-specific representations  may contain information that is irrelevant or conflicting with the tasks, which downgrades the effectiveness of learned multimodal representations. We revisit the disentanglement issue, and propose a novel triple disentanglement approach, TriDiRA, which disentangles the modality-invariant, effective modality-specific and ineffective modality-specific representations from input data. By fusing only the modality-invariant and effective modality-specific representations, TriDiRA can significantly alleviate the impact of irrelevant and conflicting information across modalities during model training. Extensive experiments conducted on four benchmark datasets demonstrate the effectiveness and generalization of our triple disentanglement, which outperforms SOTA methods.
\end{abstract}

\section{Introduction}
\label{sec:intro}

People perceive the world by collaboratively utilizing multiple senses because the multimodal sensing provides more comprehensive information from different aspects \cite{wang2022systematic,bayoudh2021survey,huang2021makes}. Recently, multimodal learning has significantly improved the performance of multimodal affective analysis (MAA, including both sentiment regression and emotion classification) due to the richer information \cite{majumder2019dialoguernn,zhou2023adaptive}. The emerging study \cite{poria2020beneath} confirmed that different modalities usually contain both consistent and complementary sentimental information, in which the complementary one can considerably boost performance.

\begin{figure}[t]
	\centering
	\includegraphics[width=88mm]{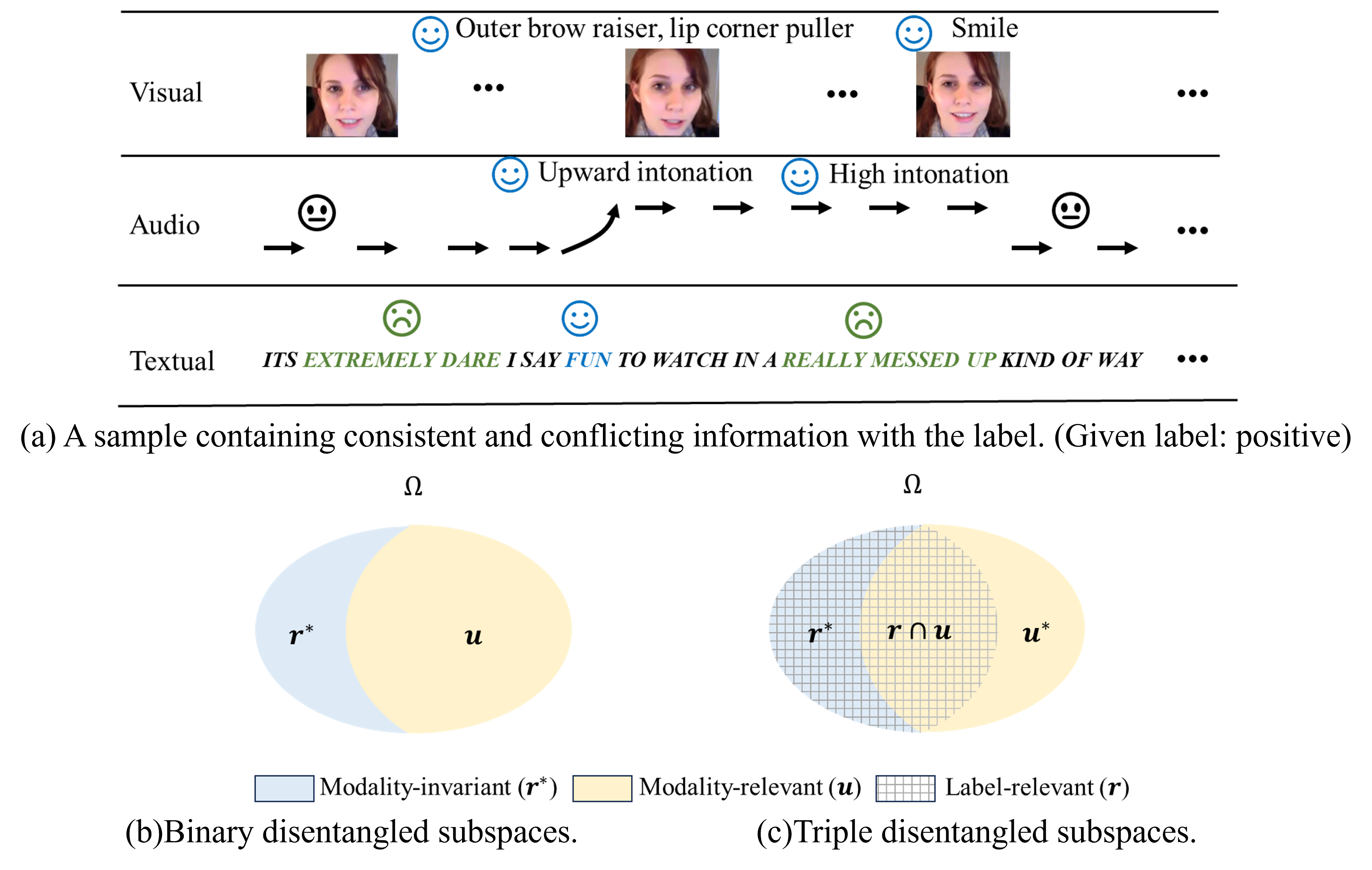}
	\caption{(a) A sample containing consistent and conflicting information among different modalities. (b) Binary disentangled subspaces of unimodality, containing modality-invariant ($r^*$) and modality-specific subspaces ($u$). (c) Triple disentangled subspaces, disentangling the modality-invariant ($r^*$), effective modality-specific ($r\cap u$), and ineffective modality-specific ($u^*$) subspaces from the label-relevant and modality-specific subspaces.}
	\label{fig:sample}
\end{figure}
Many previous approaches have developed sophisticated mechanisms to fuse the complementary information from different modalities \cite{poria2017review}. Most of them treated the information of each modality as a whole. However, due to the heterogeneity of modalities and the diverse information, they often learned unrefined or redundant multimodal representations \cite{yang2022disentangled}. Later, some studies aimed to improve the multimodal fusion by learning latent representations of modalities first, and then capturing the desirable properties from each modality. They can be broadly categorized into two groups: 1) the consistent representation learning \cite{han2021improving,mai2022hybrid,andrew2013deep}, which focuses primarily on the coherent information among modalities. It considers features from different modalities as a unified entity for learning and evaluation, leveraging constraints, such as correlation \cite{andrew2013deep}, mutual information \cite{han2021improving}, and similarity \cite{mai2022hybrid}, to guide the model towards learning primarily from modalities containing consistent information. However, such approaches inadvertently disregard the specific and complementary information of each modality. 2) the disentangled representation learning, which not only learns the consistent representations, but also learns the specific representations of each modality \cite{hazarika2020misa,li2023decoupled,yang2022disentangled}. It is a fine-grained multimodal representation learning compared to the group one. As shown in Fig. \ref{fig:sample}(b), all these methods apply binary disentanglement strategies that separates representations of each modality into modality-invariant representation $r^*$ and modality-specific representation $u$. They consider both to be label-relevant, and fuse them ($r^* \cup u$) for prediction. Our study shows that only a portion of information consistent with label semantics is significant within the modality-specific representations. For instance, as shown in Fig. \ref{fig:sample}(a), a person is recommending an interesting movie. Her facial expression and tone show that she feels the movie is `fun'. But some textual information such as the words `messed up' reflect the negative sentiment. As aforementioned methods overlook this issue, such irrelevant or conflicting information would essentially harm model training.

It is well accepted that the essential objective of MAA tasks is to obtain effective label-relevant representations only for model training. In this work, we revisit the MAA problem, and think there are two subspaces in the multimodal feature space, i.e. the label-relevant and modality-specific subspaces, named as $r$ and $u$ respectively shown in Fig. \ref{fig:sample}(c). Naturally, there exists an intersection between $r$ and $u$, which is both label- and modality-relevant, named as $r \cap u$. We can see the subspace, $r-r \cap u$, is actually the modality-invariant subspace, $r^*$, in binary disentanglement methods, while the subspace, $u^* = u- r\cap u$, holds modality-specific but label-irrelevant representations. Thus, the key point is disentangling $r \cap u$ to better fulfill MAA tasks. This reduces to a new triple disentanglement of representations: $r^*$, $r \cap u$, $u^*$.




To this end, we propose a novel triple disentanglement model, TriDiRA, which further disentangles three distinct components: \textit{modality-invariant, effective modality-specific, and ineffective modality-specific representation} from each modality. Specifically, we design two branches in TriDiRA with losses of sentiment/emotion prediction and modality classification, respectively, to obtain label-relevant ($r$) and modality-specific ($u$) representations. Then, the proposed dual-output attention module and corresponding regularizations are applied to disentangle the modality-invariant ($r^*$), effective modality-specific ($r\cap u$), and ineffective modality-specific ($u^*$)  representations from $r$ and $u$. Subsequently, the ineffective modality-specific ones are excluded by feature fusion.
Moreover, to guarantee the effectiveness of the disentanglement process, a reconstruction loss is employed to maintain the integrity of disentangled representations.

The contributions of this paper are as follows:

\begin{itemize}
	\item [1)]  We revisit the MAA problem and propose a novel triple disentanglement model, TriDiRA, which disentangles the modality-invariant and effective modality-specific representations from every modality for prediction while excluding the ineffective modality-specific representations. To the best of our knowledge, TriDiRA is the first triple disentanglement model for affective analysis tasks in the literature.
	
	\item [2)] We introduce a dual-output attention module. It can achieve a better intersection between modality-specific and label-relevant subspaces through highly dynamic interactions, which contributes to the effectiveness of triple disentanglement.
	
	\item [3)] Experiments on both sentiment regression and multi-emotion classification datasets demonstrate the effectiveness of our proposed method.
\end{itemize}

\section{Related work}
\label{sec:related}
\subsection{Multimodal representation learning}
Multimodal representation learning aims to extract effective semantic information from each modality and fuse them. Most of previous works focused on learning consistent information among different modalities. For instance, Mittal et al.  \cite{mittal2020m3er} introduced the M3ER method, which incorporates a check step utilizing Canonical Correlational Analysis (CCA)  \cite{hotelling1992relations} to effectively distinguish ineffective and effective modalities. Similarly, Sun et al.  \cite{sun2020learning} presented the ICCN model, leveraging the outer product of feature pairs in conjunction with Deep Canonical Correlation Analysis (DCCA)  \cite{andrew2013deep} to explore useful multimodal embedding features. Different from this, Han et al. \cite{han2021improving} devised MMIM, aiming at maximizing Mutual Information (MI)  \cite{tishby2015deep} at both the input and fusion levels to preserve label-relevant information during multimodal fusion. Yu et al. \cite{yu2021learning} proposed a joint training strategy for multimodal and unimodal tasks, facilitating the learning of consistency between modalities, respectively. Mai et al. \cite{mai2022hybrid}  applied intra-/inter-modal contrastive learning and semi-contrastive learning to address inter-sample and inter-class relationships while reducing the modality gap. A recent work, UniMSE\cite{hu2022unimse}, unified sentiment analysis and emotion recognition tasks, meanwhile, utilized inter-modality contrastive learning to obtain discriminative multimodal representations. Nonetheless, these methods may ignore complementary features that are private to each modality.

\begin{figure*}
	\centering
	\includegraphics[width=162mm]{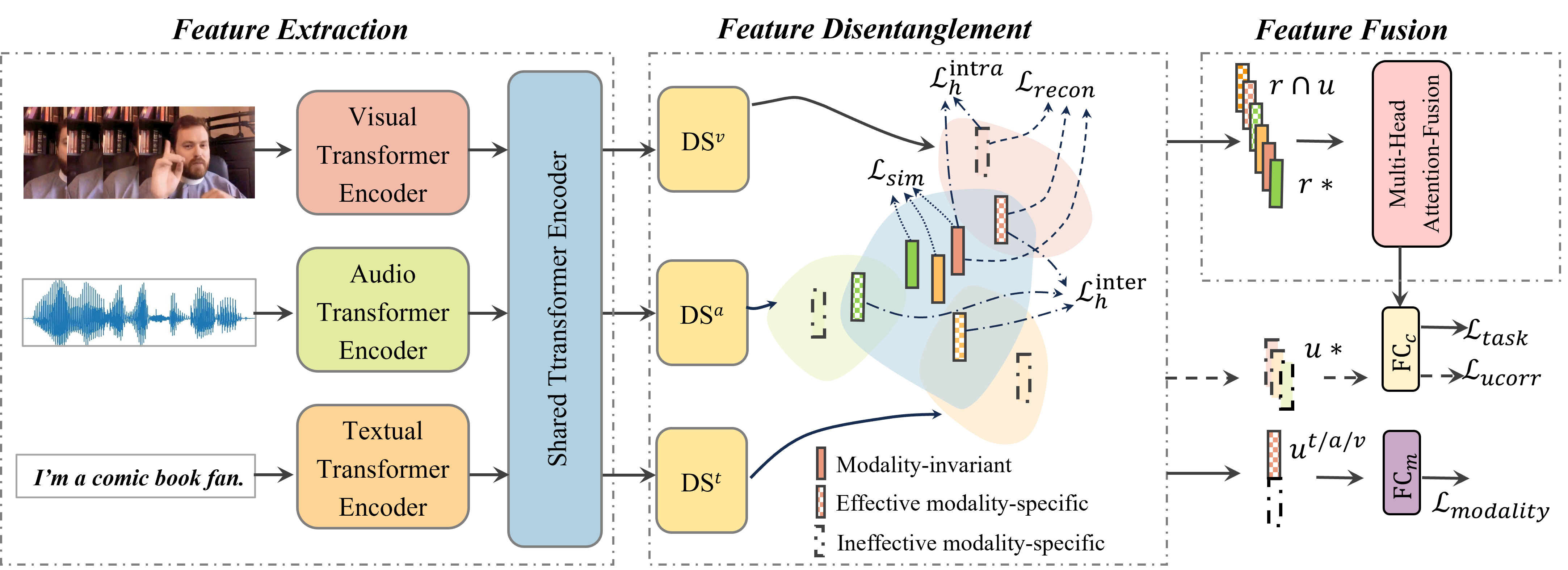}
	\caption{The flowchart of TriDiRA, includes three modules: feature extraction, feature disentanglement, and feature fusion. The feature extraction module includes three unimodal Transformer encoders and a shared Transformer encoder. The disentanglement module (DS) decomposes the unimodal features into modality-invariant representations, as well as effective and ineffective modality-specific representations. Then,  the effective representations are fused using a multi-head attention for prediction. Note that the losses $\mathcal{L}_{h}^{intra}$ and $\mathcal{L}_{recon}$ are also applied for the other two modalities.}
	\label{fig:2}
\end{figure*}

\subsection{Binary disentangled representation learning}
Recent works employ dynamic analyzing and learning features within each modality for a sample, resulting in a fine-grained multimodal representation learning approach. These methods decompose the features into two parts: modality-invariant representations shared among modalities and modality-specific representations private to each modality. Inspired by Domain Separation Network (DSN) \cite{bousmalis2016domain}, Hazarika et al. \cite{hazarika2020misa} projected the features of each modality into a modality-invariant subspace and a modality-specific subspace, which were subsequently fused using the Transformer \cite{vaswani2017attention}. Another approach, TAILOR \cite{zhang2022tailor}, also disentangled modality features into two groups and introduced a BERT-like Transformer \cite{devlin2018bert} encoder to gradually fuse these features in a granularity descent manner. Yang et al. \cite{yang2022disentangled} guided the feature disentanglement in an adversarial manner and then utilized cross-modal attention to fuse disentangled multimodal representations. In addition, Yang et al. \cite{yang2022learning} proposed to use a self-attention module to enhance the modality-specific features and a hierarchical cross-modal attention module to explore the correlations between modality-invariant features. Li et al.  \cite{li2023decoupled} proposed a disentangled multimodal distillation approach to disentangle modality-specific and modality-invariant information.  However, it is worth noting that few of the aforementioned methods take into account the presence of irrelevant or conflicting representations that may exist in modality-specific representations.

Our proposed TriDiRA can learn complementary information from the modality-specific representations while simultaneously excluding irrelevant and conflicting representations.

\section{Method}

The utterances in the multimodal affective analysis task comprise of three sequences of low-level features from text (\textit{t}), visual (\textit{v}) and audio (\textit{a}) modalities. Therefore, each sample in a training set with $B$ entities is represented as $x=\left\{ x^t,x^a,x^v \right\}$. For modality $m \in {\{t, a, v\}}$, ${x}^{m} \in  \mathbb{R}^{{\tau}^m \times {d^m} }$, ${\tau}^{m}$ represents the sequence length of the sample (i.e., the number of frames), and  ${d}^{m}$ denotes the feature dimension. The affective analysis task aims to detect the sentiment orientation or the emotion of an utterance $x$ from the label set $Y \in \mathbb{R}$. In this work, the primary objective  is to disentangle the modality-invariant representations along with the effective and ineffective modality-specific representations from the unimodal features. Subsequently, the modality-invariant and effective modality-specific representations across different modalities are fused to form a consolidated representation. It is then utilized to capture shared and complementary affective information across multiple modalities, thereby enhancing the model's performance.

Our proposed TriDiRA model comprises three modules: feature extraction, feature disentanglement, and feature fusion, illustrated in Fig. \ref{fig:2}. The details are given below.

\subsection{Feature Extraction}
Recently, Transformer-based multimodal learning has demonstrated remarkable  effectiveness in both feature extraction and fusion. However, as the number of modalities increases, the model's structure becomes increasingly intricate, resulting in a surge in the number of parameters. An emerging work  \cite{jaegle2021perceiver} shows that a unified architecture in models can handle arbitrary configurations of different modalities, which is more parameter-efficient by sharing part of the parameters among different modalities \cite{gong2022uavm}. Inspired by this, we employ a unified  module for feature extraction, as depicted in Fig. \ref{fig:2}. This module comprises three modality-specific Transformer Encoders and a shared Transformer Encoder. For the detailed structures, please refer to  the supplementary material.

Initially, the feature $x^m$ is normalized by its corresponding 1-D convolution layers, and then fed into the  ${N^m}$-layer modality-specific Transformer to obtain ${{x}^{m'}}$. Afterwards, $\{{x}^{m'}\}$ are projected to the refined embeddings $\{\hat{x}^{m}\}$ in the same feature space by the shared modality-agnostic Transformer with ${N_S}$-layers. This process not only preserves the specific characteristics of each modality, achieving implicit modality alignment, but also significantly reduces the number of parameters by nearly half compared to conventional cross-modal models. To facilitate subsequent feature disentanglement, we set the dimension of the shared Transformer Encoder to be the same as the modality-specific Transformer Encoders.

\subsection{Disentanglement module}

\begin{figure}
	\centering
	\includegraphics[width=80mm]{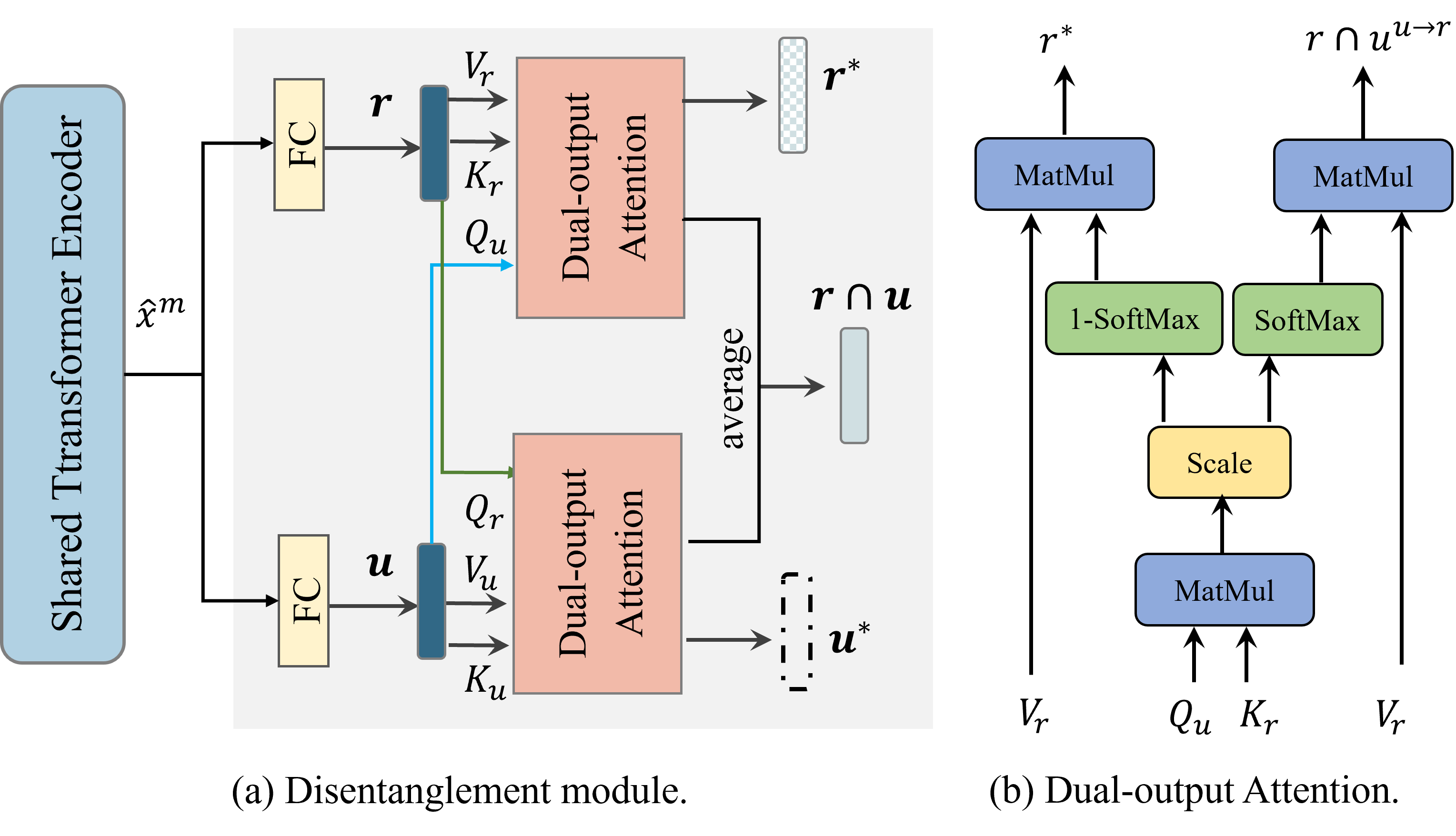}
	\caption{(a) The architecture of disentanglement module. (b) The diagram of the dual-output attention module.}
	\label{fig:disen}
\end{figure}

As stated in the introduction, it is desired to disentangle $r^*$, $r\cap u$ and $u^*$ from $x^m$. We expect that $r^*$ is label-relevant and modality-invariant, $r\cap u$ is label-relevant and modality-specific, and $u^*$ is modality-specific while label-irrelevant. To this end, we introduce a novel triple disentanglement module for the first time. Its architecture is illustrated in Fig. \ref{fig:disen}(a). The output of shared Transformer encoder $\hat{x}^m$ is firstly processed by two fully-connected layers (FCs), which produce representations $r$ and $u$, respectively.

To further obtain the modality-invariant representation $r^*$ and effective modality-specific representation $r \cap u$, we design a dual-output attention module as shown in Fig. \ref{fig:disen}(b). For the representations $u$ and $r$, the query $Q_u$  is generated from $u$, $Q_u=uW^{Q_u}$ and $K_r$, $V_r$  are generated from $r$, $K_r=rW^{K_r}$, $V_r=rW^{V_r}$. Then two components can be disentangled from the representation $r$,

\begin{equation}
	\begin{split}
		{r \cap u}^{u \rightarrow r}= Attention^{u \rightarrow r} (Q_u,K_r,V_r)\\
		=softmax(\frac{Q_u {K_r}^T}{\sqrt{d_k}}) V_r,
	\end{split}
\end{equation}

\begin{equation}
	r^*=(1-softmax(\frac{Q_u {K_r}^T}{\sqrt{d_k}}))V_r.
\end{equation}

Similarly,  $r \cap u^{r \rightarrow u} $and $u^*$ can be disentangled from $u$.

Then, $r \cap u$ is computed by

\begin{equation}
	r \cap u=\frac{{r \cap u}^{u \rightarrow r}+r \cap u^{r \rightarrow u}}{2}.
\end{equation}

These disentangled representations are obtained by a joint optimization of the task losses, similarity losses and independent losses in sections 3.2.1, 3.2.2, and 3.2.3. Note that $r$ and $u$ are going to be label-relevant and modality-specific, respectively, with the progress of the optimization.

\subsubsection{Task Losses}

As $r^*$, $u^*$ and $r \cap u$ are either label-relevant, modality-specific, or both-relevant, we devise two task constraints for each of them to optimize the two branches of the disentanglement module in Fig. \ref{fig:disen}(a), respectively.

Due to both $r^*$ and $r \cap u$ (i.e. $r$) of all modalities are relevant to the label semantics, they are constrained by the regression/classification loss of the affective analysis task. Specifically, they are stacked and then fused by a multi-head attention, as shown in the feature fusion module of Fig. \ref{fig:2}. Afterwards, a $FC_c$ layer gives the predictions $\hat{y}_i$. The difference between $\hat{y}_i$ and the ground truth $y_i$ is minimized by the following loss,

\vspace{-0.1 in}
\begin{equation}
	\begin{split}
		\mathcal{L}_{task}= \frac{1}{B} \sum_{i=1}^B {\parallel y_i-\hat{y}_i  \parallel}^2_2 \quad (regression)\\
		= -\frac{1}{B} \sum_{i=1}^B { y_i \cdot log \hat{y}_i } \quad (classification).
	\end{split}	
\end{equation}

By contrast, both $u^*$ and $r \cap u$ (i.e. $u$) of all modalities are modality-specific, they are constrained by a modality classification task. In particular, $u^*$ and $r \cap u$ of each modality are concatenated, and then fed into a $FC_m$ layer. The output is optimized by the following loss,

\begin{equation}
	\mathcal{L}_{modality}^m = -\frac{1}{B} \sum_{i=1}^B {\mathbb{I} (m) \cdot {log D(u_i^{m})}},
\end{equation}
where, $\mathbb{I}(m)$ is a one-hot modality label, and $D$ denotes the modality discriminator.

Moreover, $u^*$ should be label independent, we constrain it by minimizing the following loss,

\begin{equation}
	\begin{split}
		\mathcal{L}_{ucorr}=\frac{cov(\tilde{Y},Y)}{\sigma_{\tilde{Y}} \sigma_{Y}} \quad (regression),\\
		= \frac{1}{B} \sum_{i=1}^B { y_i \cdot log \tilde{y}_i } \quad (classification).
	\end{split}
\end{equation}
where, $\tilde{y_i}$ is the prediction of the $FC_c$ layer by feeding the ineffective modality-specific representation $u_i^*$. $\tilde{Y}$ is the set of $\tilde{y_i}$ and $Y$ denotes the set of $y_i$. The $cov(\tilde{Y},Y)$ refers to covariance and $\sigma$ refers to the variance.

\subsubsection{Similarity Loss}

As the modality-invariant representations $\{r^{*,m}\}$ of modalities share common motives and goals of the speaker, they should be mapped to a shared subspace, in which capturing the underlying commonalities and aligned features correlated with labels. To this end, we employ the Central Moment Discrepancy (CMD) \cite{zellinger2017central} to align them, which can minimize the heterogeneity among modalities. CMD distance decreases as two distributions become more similar. In this study, let $Z$ and $W$ be two feature distributions of two modalities, their similarity is computed by an empirical estimate of the CMD metric,

\vspace{-0.1 in}
\begin{equation}
	\begin{split}
		CMD_K(Z,W) &= \frac{1}{|b-a|}\parallel E(Z)-E(W)\parallel_2 \\
		&+\sum_{k=2}^K \frac{1}{|b-a|^k}\parallel C_k(Z)-C_k(W)\parallel_2,
	\end{split}
\end{equation}
where, $a$ and $b$ are the minimal and maximal values of features,  $E(Z)=\frac{1}{|Z|}\sum_{z \in Z} z$ is the empirical expectation vector of $Z$ and $C_k(Z) = E((z-E(Z))^k)$  is the central moment vector of order $k$.

Thus, we have a loss for maximizing the similarity between the invariant representations of every pair of modalities,

\vspace{-0.1 in}
\begin{equation}
	\mathcal{L}_{sim}=\frac{1}{3}\sum_{(m_1,m_2)\in \\ {(t,a),(t,v),\\(a,v)}} CMD_K(r^{*,m_1},r^{*,m_2}).
\end{equation}

\subsubsection{Independence Losses}

As each modality-specific representation $(r \cap u)^{m}$ is uncorrelated to those of other modalities, they should be independent. Moreover, $r^{*,m}$ and $u^{*,m}$ should hold exclusive characteristics of the $m$ modality. To this end, we employ the Hilbert-Schmidt Independence Criterion (HSIC) loss \cite{greenfeld2020robust}, which ensures the modality-specific and modality-invariant representations to capture distinctive information of an utterance. HSIC loss is defined as

\begin{equation}
	\mathcal{L}_h^{Z,W}= \frac{1}{(B-1)^2}Tr(\Phi (Z,Z)J\Phi(W,W)J),
\end{equation}
where $Tr$ is the trace of a matrix, $J=I-\frac{1}{B}$, $I$ is an identify matrix, $\Phi(z,w)=exp(\frac{-\parallel z \cdot w ^T\parallel}{2 \sigma^2})$.

Thus, for the independence across modalities, we devise an inter-independent loss for every pair of modalities,
\begin{equation}
	\mathcal{L}_h^{inter} = \frac{1}{3}\sum_{(m_1,m_2)\in \\ {(t,a),(t,v),\\(a,v)}}\mathcal{L}_h^{(r \cap u)^{m_1},(r \cap u)^{m_2}}.
\end{equation}

Note that $u^{*,m}$ is optimized by the modality classification loss in Eq. 6. The inter-independence between them has been guaranteed.

Meanwhile, for the exclusivity  within each modality, we enforce an intra-exclusive constraint between $r^{*,m}$ and $u^{*,m}$,

\begin{equation}
	\mathcal{L}_h^{intra} = \mathcal{L}_h^{r^{*,m},u^{*,m}}.
\end{equation}

\subsubsection{Reconstruction Loss}

In addition, we apply a reconstruction task to ensure that the disentangled representations $r^{*,m}$, $(r \cap u)^m$ and $u^{*,m}$ can capture the essential features without losing any informative details of the utterance. The task aims to recover the input features, $\hat{x}^m$, of the disentanglement module in Fig. \ref{fig:disen}(b).

\begin{equation}
	\mathcal{L}^m_{recon}= \frac{1}{B} \sum_{i=1}^B {\parallel \hat{x}^{m'} -\hat{x}^m \parallel}^2_2,
\end{equation}
where $\hat{x}^{m'}$ is generated  by a decoder $R(r^{*,m}, (r \cap u)^m, u^{*,m})$.

Finally, the effectiveness of feature disentanglement is jointly optimized by all the aforementioned losses with different trade-off weights.

\section{Experiments}
\subsection{Datasets}
\begin{table*}[]
	\centering
	\caption{Performance comparison among TriDiRA and seven SOTA methods on CMU-MOSI and CMU-MOSEI. In Acc-2 and F1-Score, the left of the “/” is calculated as “negative/non-negative” and the right is calculated as “negative/positive”. The best results are in bold and the second-best results are underlined.  The results in the upper half of the table are from the original papers. The results in the lower half are re-implemented using our features and settings for a fair comparison. $\dag$ refers to binary disentanglement learning methods.}
	\label{tab:comparison}
	\resizebox{180mm}{28mm}{
		\begin{tabular}{l|ccccc|ccccc}
			\hline
			\multirow{2}{*}{Methods} & \multicolumn{5}{c|}{MOSI}    & \multicolumn{5}{c}{MOSEI}   \\
			& MAE($\downarrow$) & Corr($\uparrow$) & Acc-2($\uparrow$) & F1-Score($\uparrow$) & Acc-7($\uparrow$)
			& MAE($\downarrow$) & Corr($\uparrow$) & Acc-2($\uparrow$) & F1-Score($\uparrow$) & Acc-7($\uparrow$) \\ \hline
			MMIM \cite{han2021improving}  & 0.700         & 0.800        & 84.14/86.06 & 84.00/85.98 & 46.65 & 0.526 & 0.772 & 82.24/85.97 & 82.66.85.94 & 54.24 \\
			HyCon \cite{mai2022hybrid} & 0.664         & 0.832        & -/86.4      & -/86.4      & 48.3  & 0.590 & 0.792 & -/86.5      & -/86.4      & 53.4  \\
			UniMSE \cite{hu2022unimse} & 0.691         & 0.809        & 85.85/86.9  & 85.83/86.42 & 48.68 & 0.523 & 0.773 & 85.86/87.50 & 85.79/87.46 & 54.39 \\
			MISA \cite{hazarika2020misa} $\dag$ & 0.783         & 0.761        & 81.8/83.4   & 81.7/83.6   & 42.3  & 0.555 & 0.756 & 83.6/85.5   & 83.8/85.3   & 52.2  \\
			FDMER \cite{yang2022disentangled} $\dag$ & 0.724         & 0.788        & -/84.6      & -/84.7      & 44.1  & 0.536 & 0.773 & -/86.1      & -/85.5      & 54.1  \\
			MFSA \cite{yang2022learning} $\dag$ & 0.856         & 0.722        & -/83.3      & -/83.7      & 41.4  & 0.574 & 0.734 & -/83.8      & -/83.6      & 53.2  \\
			DMD  \cite{li2023decoupled} $\dag$ & -             & -            & -/83.5      & -/83.5      & 41.9  & -     & -     & -/84.8      & -/84.7      & 54.6  \\\hline
			MMIM\cite{han2021improving}   & \underline{0.706}         & \underline{0.798}        & 83.18/\underline{84.62} & 83.12/\underline{84.46} & 47.23 & 0.540 & 0.760 & \underline{82.46}/84.98 & 82.79/84.89 & \underline{53.44} \\
			HyCon \cite{mai2022hybrid} &    0.709    &    0.792      &   82.40/83.20     &    81.70/83.10   & \underline{47.80} &  0.593  &  0.768  &    82.16/85.42     &     81.96/85.02    &  46.72  \\
			MISA \cite{hazarika2020misa} $\dag$  & 0.749         & 0.785        & 81.20/82.93 & 81.19/82.98 & 43.73 & 0.549 & 0.758 & 80.30/84.84 & 80.96/84.90 & 52.80 \\
			FDMER \cite{hazarika2020misa}	$\dag$ &  0.721     &     0.790   &   82.65/84.30   &   82.60/84.29    &  44.31  &  0.539   &   \underline{0.769} &    81.00/85.06   &   81.59\underline{/85.09}   &  52.16  \\
			MFSA \cite{yang2022learning} $\dag$  &  0.851     &    0.734       &     80.20/81.70     &    80.12/81.90    & 42.30   &  0.577   &  0.741  &   80.02/82.10      &   80.01/81.70    &  52.70     \\
			DMD   \cite{li2023decoupled} $\dag$  & 0.720         & 0.793        & \underline{83.22}/84.30 & \underline{83.13}/84.21 & 46.36 & \underline{0.536} & \underline{0.769} & \underline{82.46/85.50} & \underline{82.88}/85.04 & 52.35 \\
			TriDiRA  & \textbf{0.673}         & \textbf{0.813}        & \textbf{83.24/84.76} & \textbf{83.15/84.72} & \textbf{50.00} & \textbf{0.529} & \textbf{0.775} & \textbf{84.85/85.77} & \textbf{84.80/85.47} & \textbf{53.49} \\ \hline
		\end{tabular}
	}
\end{table*}
To compare with the existing disentangled methods \cite{hazarika2020misa,yang2022disentangled,yang2022learning,li2023decoupled}, we follow their experimental protocol and test all methods on three benchmark datasets CMU-MOSI \cite{zadeh2016mosi}, CMU-MOSEI \cite{zadeh2018multimodal} and UR-FUNNY \cite{hasan2019ur}, which provide multimodal signals (text, visual and audio) for each utterance. As these datasets utilize the sentiment polarities or intensities for the regression/classification task, we include an additional dataset, MELD \cite{poria2018meld} with six emotional classes, to evaluate the generalization of all methods. Moreover, unlike the data collected in the laboratory (e.g. IEMOCAP), the utterances in all above datasets were collected in the wild which are the main target of our test.

\textbf{CMU-MOSI} contains 2199 utterance video segments sliced from 93 videos in which 89 individuals express their opinions on interesting topics. Each segment is manually annotated with a sentiment value ranging from -3 to +3, indicating the polarity (by positive/negative) and relative strength (by absolute value) of expressed sentiment.

\textbf{CMU-MOSEI} upgrades CMU-MOSI by expanding the size of the dataset. It contains 22856 annotated video segments (utterances), from 5000 videos, 1000 individuals and 250 different topics.

\textbf{UR-FUNNY} provides 16514 samples of multimodal utterances from TED talks with diverse topics and speakers. Each utterance is labeled with a binary label as humor/non-humor.

\textbf{MELD} contains 13,707 video clips of multi-party conversations, with labels following Ekman’s six universal emotions, including joy, sadness, fear, anger, surprise and disgust.

\subsection{Evaluation metrics}
Following works \cite{han2021improving,hazarika2020misa,yang2022disentangled}, evaluations are conducted on two tasks: classification and regression. For CMU-MOSI and CMU-MOSEI datasets, binary accuracy (Acc-2), F1-Score and seven-class accuracy (Acc-7) are reported on the classification task. Note that Acc-2 and F1-Score are calculated in two ways: negative/non-negative (include zero) and negative/positive (exclude zero). We also report Mean Absolute Error (MAE) and Pearson Correlation (Corr) on the regression task. For UR-FUNNY dataset, binary accuracy (Acc-2) and F1-Score are reported. For MELD dataset, six-class accuracy (Acc-6) is reported. Except for MAE, higher values denote better performance for all metrics.

\subsection{Experimental settings}
For textual features, the BERT-base-uncased pre-trained model  \cite{devlin2018bert} is employed on all datasets. On CMU-MOSI, CMU-MOSEI and MELD datasets, librosa \cite{mcfee2015librosa} is applied to extract the Mel-spectrogram to obtain the acoustic features, and the pre-trained EffecientNet \cite{tan2019efficientnet} is used for visual features.  UR-FUNNY applies COVAREP \cite{degottex2014covarep} for acoustic features and OpenFace \cite{baltrusaitis2018openface} for facial expression features.  The multimodal features of UR-FUNNY are word-aligned, while the other three use word-unaligned features. All methods are trained and tested with one RTX 3090 GPU. Each test was performed five times, and the average result is reported. More parameter configurations and experimental details are shown in the supplementary materials.



\subsection{Comparison with SOTA methods}
\subsubsection{Competing methods}

As TriDiRA is the first triple disentangled method, the performance comparison is mainly conducted with the binary disentanglement methods \cite{hazarika2020misa,zhang2022tailor,yang2022disentangled,yang2022learning,li2023decoupled}. We firstly checked their reported performances in the literature, and then selected the top three (FDMER \cite{yang2022disentangled}, DMD \cite{li2023decoupled} and MFSA \cite{yang2022learning}) and the representative (MISA \cite{hazarika2020misa}) for comparison. We also examined multimodal representation learning methods \cite{mittal2020m3er,sun2020learning,yu2021learning,han2021improving,mai2022hybrid,hu2022unimse}. The top three (MMIM \cite{han2021improving}, UniMSE \cite{hu2022unimse} and HyCon \cite{mai2022hybrid}) were included into the comparison as well.

\subsubsection{Results and analysis}

\textbf{4.4.2.1. Sentiment analysis}

The evaluations on MOSI and MOSEI are listed in Table \ref{tab:comparison}. It can be observed that TriDiRA outperforms all re-implemented competing methods on most of the metrics (regression and classification combined).

Firstly, TriDiRA outperforms all the binary disentangled learning methods. The underlying reason is that the modality-specific representation $u$ disentangled by the binary disentangled methods contain some ineffective information, please refer to the comparative validation of $u$ and $r \cap u$ in Section 4.5. Different from binary disentanglement, the substantial performance gap between disentangled $u^*$ and $r^*$ proves that TriDiRA does eliminate those label-irrelevant representations, resulting in enhanced performance.

As for the consistent representation learning methods, both MMIM and HyCon focus on minimizing the modality gap, which can effectively improve the multimodal fusion and prediction, but may ignore some modality-specific representations. TriDiRA can further utilize these representations to obtain complementary information, thus achieving a better performance.

An emerging work, UniMSE, unifies four heterogeneous datasets via generating universal labels. It considerably boosts the representation learning by largely expending the training data, and achieves a SOTA record. In contrast, TriDiRA relies solely on the given dataset and reaches a comparable result thanks to eliminating ineffective representation during learning.

\begin{table}[]
	\caption{Performance comparison among TriDiRA and seven SOTA methods on UR-FUNNY and MELD. \protect\footnotemark}
	\label{tab:ur}
	\centering
	\resizebox{67mm}{18mm}{
		\begin{tabular}{lc|cc}
			\hline
			{ } & \multicolumn{1}{c|}{UR-FUNNY} & \multicolumn{2}{c}{MELD} \\
			& ACC-2($\uparrow$)              & ACC-6($\uparrow$)      & F1-Score($\uparrow$)       \\
			\hline
			UniMSE \cite{hu2022unimse}&   -                        & 65.09       & 65.51      \\
			MISA \cite{hazarika2020misa} $\dag$  & 70.61                     & -           & -          \\
			FDMER \cite{yang2022disentangled} $\dag$ & 71.87                   & -           & -          \\\hline
			HyCon  \cite{mai2022hybrid} &   68.20    &   61.24      &   60.72    \\
			MISA \cite{hazarika2020misa} $\dag$ &     70.06            &  $\star$   &   $\star$    \\
			FDMER  \cite{hazarika2020misa} $\dag$&    \underline{71.09}             &   \underline{63.79}          &    \underline{61.59}        \\
			MFSA  \cite{yang2022learning} $\dag$&     70.30                   &   61.23     &   60.92     \\
			DMD  \cite{yang2022learning} $\dag$&     70.01            &  $\star$    &    $\star$   \\
			TriDiRA   &   \textbf{72.58}        &     \textbf{65.56}     &    \textbf{63.44}       \\ \hline
	\end{tabular}}
\end{table}

\textbf{4.4.2.2. Humor detection}

Since humor data often contain inconsistent information across modalities, we specifically validate the effectiveness of TriDiRA on UR-FUNNY, which is a binary classification dataset. The results in Table \ref{tab:ur} show a similar trend as the results of ACC-2 in Table \ref{tab:comparison},  and TriDiRA also achieves the best performance.

\textbf{4.4.2.3. Emotion classification}

To test the robustness of TriDiRA and other methods, a comparison is  also conducted on the more challenging emotion classification dataset, MELD. The results in the Table \ref{tab:ur} demonstrate TriDiRA outperforms other methods by capturing effective emotional information. In short, the superiority of TriDiRA on both sentiment regression and multi-emotion classification tasks indicates its notable generalization ability.

\footnotetext{$\star$ denotes that the method is unable to be reproduced on the dataset. MISA and DMD were not validated on multi-emotion data in their papers. Meanwhile, our implementations using their source codes were failed to converge on the MELD dataset. MMIM was designed for regression tasks. Its code does not work on MELD either.}
\subsection{Disclosure of dientangled representations}

\begin{table}[]
	\centering
	\caption{Comparison between representations disentangled by triple and binary methods on sentiment prediction using a third-party evaluator (Rep. stands for representations).}
	\label{tab:third}
	\resizebox{84mm}{10mm}{
		\begin{tabular}{ccccccc}
			\hline
			Methods & Rep. & MAE($\downarrow$) & Corr($\uparrow$) & Acc-2($\uparrow$) & F1-Score($\uparrow$) & Acc-7($\uparrow$) \\ \hline
			\multirow{3}{*}{\makecell{(Triple)\\ TriDiRA}} & $r^*$ & 0.685 & 0.813  & 83.38/84.91  & 83.32/84.89 & 46.79 \\
			& $r \cap u$ & 0.687 & 0.813  & 83.09/84.60  & 83.03/84.59 & 46.65 \\
			& $u^*$ & 1.98  & -0.060 & 55.248/57.77 & 39.32/42.31 & 22.74 \\ \hline
			\multirow{2}{*}{\makecell{(Binary)\\ DMD}}     & $r^*_{DMD}$ & 0.715 & 0.791  & 82.07/84.15  & 81.93/84.08 & 45.92 \\
			& $u_{DMD}$   & 0.714 & 0.791  & 81.78/83.84  & 81.65/83.79 & 46.21 \\ \hline
	\end{tabular}}
\end{table}

\begin{table}[]
	\centering
	\caption{Comparison between representations disentangled by triple and binary methods on modality classification.}
	\label{tab:modality}
	\resizebox{64mm}{7mm}{
		\begin{tabular}{lccc|cc}
			\hline
			Methods & \multicolumn{3}{c|}{TriDiRA (Triple)} & \multicolumn{2}{c}{DMD (Binary)} \\
			Rep.    & $r^*$   & $r \cap u$     &  $u^*$  &   $r^*_{DMD}$        &  $u_{DMD}$               \\ 	\hline
			Acc     & 47.28      & 100.00     & 100.00     & 33.19          & 100.00          \\
			F1-score      & 38.28      & 100.00     & 100.00     & 18.99          & 100.00     \\ \hline
	\end{tabular}}
\end{table}


The aforementioned comparisons with SOTA methods show the superiority of TriDiRA, and it is worth having an insight into the underlying mechanism. Thus we evaluate the effectiveness of the disentangled representations $r^*$, $u^*$ and $r \cap u$, and compare them with the $r^*_{DMD}$ and $u_{DMD}$  disentangled by the latest SOTA binary disentanglement method, DMD \cite{li2023decoupled}, on the commonly accepted dataset MOSI. To this end, we introduce two third-party evaluators, one for sentiment prediction and another for modality classification.  Specifically, we first use the trained TriDiRA and DMD models to extract representations from the training set of MOSI. Then, an evaluator composed of two FC layers is trained on these representations for sentiment prediction and another analogous evaluator is trained for modality classification. Afterwards, two evaluators test all disentangled representations of both methods on the test set, respectively.

As shown in Table \ref{tab:third}, the disentangled representations $r^*_{DMD}$ and $u_{DMD}$  of DMD achieve impressive results. In contrast, TriDiRA significantly improves the effective representations $r^*$ and $r \cap u$, resulting in an approximately 3\% increase in MAE and a 2\% increase in correlation. From the perspective of the third-party evaluator, the representations disentangled by TriDiRA contain more effective information. A more important observation is the pronounced performance gap between the ineffective modality-specific representation $u^*$ and the effective representations $r^*$ and $r \cap u$. The extremely low correlation (-0.06) and classification accuracies suggest that $u^*$ is nearly irrelevant to the label, which proves that TriDiRA successfully disentangles effective and ineffective representations. The feature visualization in Fig. \ref{fig:visual} also supports this conclusion. Furthermore, the results of modality classification in Table \ref{tab:modality} demonstrate that disentangled representations $r \cap u$, $u^*$ and $u_{DMD}$  reach 100\% accuracies, nevertheless,  $r^*$ and $r^*_{DMD}$ reach extremely low performance. The results confirm that both methods effectively disentangle modality-invariant and modality-specific representations, respectively. Combining the results of Table \ref{tab:third} and \ref{tab:modality}, $u^*$ is indeed the ineffective modality-specific representation, and $r \cap u$ is both label-relevant and modality-specific. We believe that TriDiRA's superiority comes from disentangling  $u^*$ from $u$, thereby enhancing the quality of effective information in $r^*$ and $r \cap u$. By contrast, DMD may include some ineffective information in $u_{DMD}$, which results in slightly lower performance.

Figure \ref{fig:visual} visualizes the learned representations without or with the disentanglement regularizations of both methods. One can see these regulations pull the modality-invariant representations closer in the same subspace, while our TriDiRA can further completely disentangle the effective and ineffective modality-specific representations.

\begin{figure}
	\centering
	\includegraphics[width=80mm]{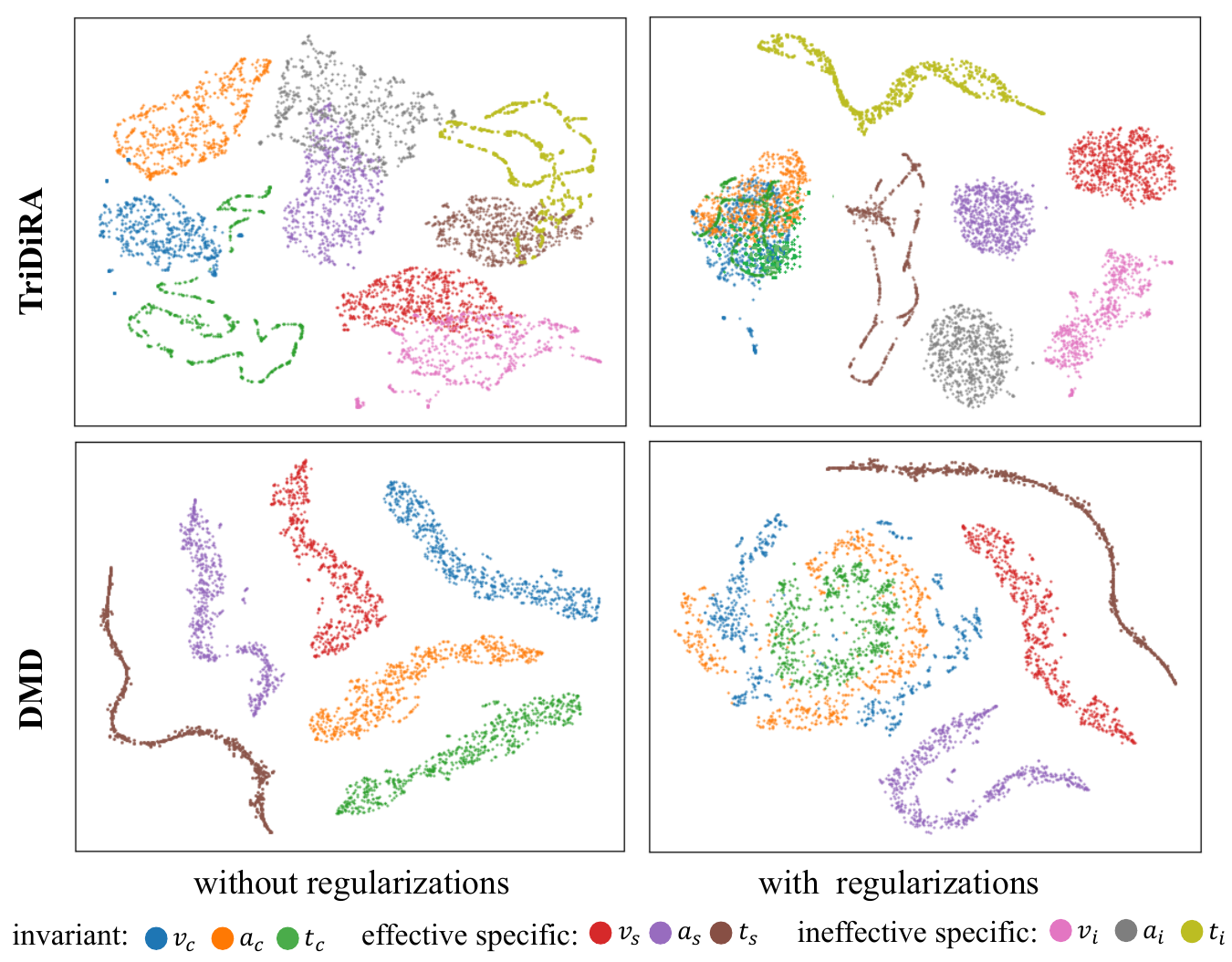}
	\caption{The t-SNE visualization of the modality-invariant and modality-specific (effective and ineffective) subspaces of TriDiRA and DMD on the test  set of MOSI.}
	\label{fig:visual}
\end{figure}

\subsection{Ablation study}

As illustrated in Fig. \ref{fig:2}, TriDiRA consists of three modules: feature extraction, feature disentanglement, and feature fusion. There are five loss functions employed for model optimization. We evaluate each module and loss function on the MOSI dataset, and report their effectiveness in Table \ref{tab:ablation}.

\textbf{Importance of modules:} We set up a baseline to test the impact of each key module. It is composed of the three modality-specific Transformer Encoders and the multi-head self-attention Transformer as in Fig. \ref{fig:2}, because both of them are widely adopted approaches for feature extraction and feature fusion. Then, the Disentangled module (DS) and Shared Transformer (ST) module are separately plugged into the baseline for evaluation. We can observe DS module plays a significant role in improving the model’s performance. ST module can align heterogeneous multimodal features, thereby enhancing the model’s performance, particularly on the regression task.  DS module not only aligns the effective representations among different modalities, but also captures complementary information. Combining these two modules will enhance the ability of DS module to extract effective representations.

\begin{table}[]
	\caption{Ablation study of two modules and five losses on MOSI.}
	\label{tab:ablation}
	\resizebox{0.5\textwidth}{!}{
		\begin{tabular}{l|ccccc}
			\hline
			Model       & MAE($\downarrow$) & Corr($\uparrow$) & Acc-2($\uparrow$) & F1-Score($\uparrow$) & Acc-7($\uparrow$) \\  \hline
			\multicolumn{6}{c}{Importance of Module}                            \\ \hline
			Baseline     & 0.711         & 0.798        & 81.92/84.30 & 81.72/84.19 & 46.50 \\
			+ST          & 0.704         & 0.803        & 82.07/84.45 & 81.76/84.25 & 46.21 \\
			+DS          & 0.688         & 0.804        & 83.07/84.47 & 83.39/85.29 & 47.08 \\ \hline
			\multicolumn{6}{c}{Importance   of regularizations}                                      \\ \hline
			w/o $\mathcal{L}_{sim}$   & 0.686         & 0.808        & 82.94/84.21 & 82.76/83.29 & 47.81 \\
			w/o $\mathcal{L}_{ucorr}$ & 0.688         & 0.810        & 82.65/84.10 & 82.47/84.02 & 46.21 \\
			w/o $\mathcal{L}_{recon}$ & 0.694         & 0.803        & 81.65/82.60 & 81.54/82.56 & 45.63 \\
			w/o $\mathcal{L}_{modality}$ & 0.680         & 0.807        & 82.65/84.67 & 82.47/84.64 & 48.25 \\
			w/o $\mathcal{L}_{h}$     & 0.680         & 0.809        & 82.51/84.62 & 82.25/84.58 & 48.40 \\
			\hline
			\textbf{TriDiRA}    & \textbf{0.673}         & \textbf{0.813}        & \textbf{83.24/84.76} & \textbf{83.15/84.72} & \textbf{50.00}  \\
			\hline
			
		\end{tabular}
	}
\end{table}

\textbf{Importance of regularizations:} To quantitatively verify the impact of five losses, we re-train TriDiRA by ablating one loss at a time, and report the results in Table \ref{tab:ablation}. As one can see, the joint optimization with all losses achieves the best performance. In a closer look, TriDiRA is rather sensitive to the $\mathcal{L}_{sim}$, $\mathcal{L}_{ucorr}$ and $\mathcal{L}_{recon}$. $\mathcal{L}_{sim}$ is responsible for pulling the representations of different modalities into a subspace, ensuring the modality invariance. $\mathcal{L}_{ucorr}$ constrains the independence of $u^*$ from labels, enabling TriDiRA to effectively disentangle ineffective information from $u$ and enhance the quality of effective information in $r \cap u$. $\mathcal{L}_{recon}$ ensures the information integrity of three disentangled representations. It is critical because all essential information of each modality is expected to be preserved after disentangling. We also see, $\mathcal{L}_{modality}$ and $\mathcal{L}_{h}$ are less dependent on TriDiRA. One possibility is that three modalities are inherent heterogenetic, which is proved by visualization without regulations in Fig. \ref{fig:visual}. But, $\mathcal{L}_{modality}$ and $\mathcal{L}_{h}$ can further enhance the independence of the disentangled representation.

\section{Conclusion}
We present TriDiRA - a novel triple disentangled representation learning method to prevent the model from being perturbed by irrelevant and conflicting information in the modality-specific representations. Despite appearing to be an upgrade of binary disentanglement learning, TriDiRA addresses a core issue overlooked by binary approaches by examining the nature of MAA tasks. Thus, it demonstrates significant gains and better generalization over SOTA approaches on both sentiment regression and emotion classification tasks. Explorative analysis by two third-party evaluators reveals the desirable trait, which is that the quality of effective representation is enhanced by excluding the label-irrelevant representations.
In future work, the incorporation of similarity and independence loss modeling allows a diverse array of regularization alternatives. Consequently, we plan to analyze other options to further enhance the triple disentanglement, especially, the quality of modality-invariant representations.

\bibliographystyle{ieeenat_fullname}
\bibliography{main}
\appendix
\newpage
\setcounter{page}{1}
\renewcommand\thesection{\Alph{section}}

\twocolumn[{%
	\renewcommand\twocolumn[1][]{#1}%
	\maketitle
	\begin{center}
		\centering
		\captionsetup{type=figure}
		\includegraphics[width=120mm]{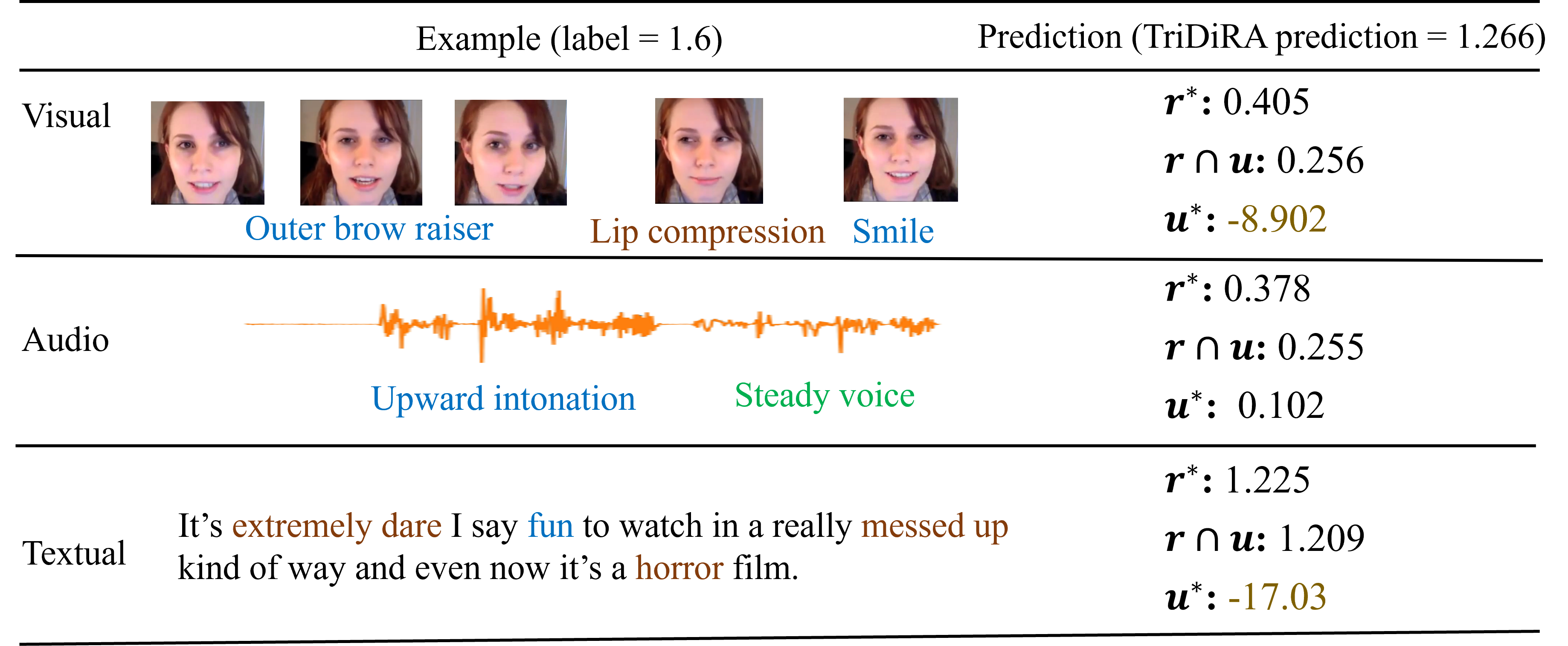}
		\captionof{figure}{An instance with label 1.6, whose modalities contain conflicting or irrelevant information. And, the performance of representations disentangled by TriDiRA. Blue denotes positive, brown denotes negative, and green denotes irrelevant to sentiment prediction.}
		\label{fig:sample1}
	\end{center}%
	
	\begin{center}
		\centering
		\captionsetup{type=figure}
		\includegraphics[width=120mm]{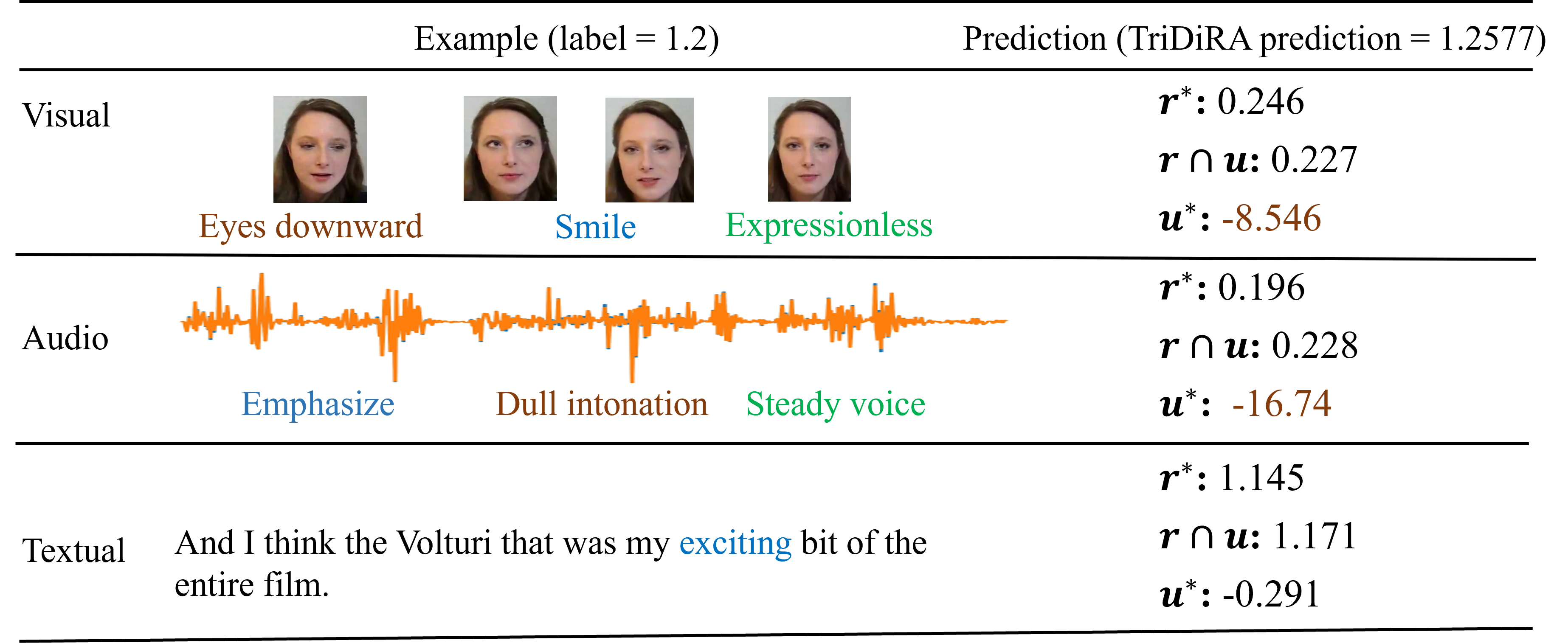}
		\captionof{figure}{Another instance with label 1.2, whose modalities contain conflicting or irrelevant information. And, the performance of representations disentangled by TriDiRA. }
		\label{fig:sample2}
	\end{center}%
}]
\maketitle
\section{Overview}
In this supplementary material, we present more experimental results and analysis in Sec. \ref{sec:sample} to further prove the effectiveness of the proposed TriDiRA. Sec. \ref{fig:visual} visualizes the attention process in the multi-head attention module. In Sec. \ref{sec:complexity}, the complexity analysis of TriDiRA compared with other methods is conducted. Sec. \ref{sec:setting} shows the detailed  architecture as well as the hyperparameter settings in TriDiRA. Sec. \ref{sec:loss} exhibits the convergence trends of different regularizations on the training set.

\section{Case study}
\label{sec:sample}
Sec. 4.5 of the main paper demonstrates the effectiveness of triple disentanglement at the statistical level; in this section, the validity of TriDiRA is verified at the instance level.  We selected two examples from CMU-MOSI and separately utilized the desentangled representations $r^*$, $r \cap u$, and $u^*$ of each modality for sentiment prediction, as shown in Fig. \ref{fig:sample1} and Fig. \ref{fig:sample2}.

In Fig. \ref{fig:sample1}, each modality contains information consistent with the label, such as the smiling in vision, upward intonation in audio, and the word `fun' in text. However, text and vision  also include conflicting information like `messed up' and lip compression, respectively. And the steady voice in audio is irrelevant to sentiment prediction. We can see that predictions of both $r^*$ and $r \cap u$ from all modalities are correlated with the label, particularly those of text. Meanwhile, text contains information conflicting with the label, and the prediction of $u^{*,t}$ deviates significantly from the label. Additionally, since the audio modality contains more information that is irrelevant with sentiment prediction, the prediction of $u^{*,a}$ is close to  0. Combining both $r^*$ and $r \cap u$ can lead to a more accurate prediction.
Similar results of another instance are illustrated in Fig. \ref{fig:sample2}.

These two instances qualitatively demonstrate that $u^*$ disentangled by TriDiRA indeed contains information conflicting/irrelevant to the label. Thus, by disentangling $u^*$ from $u$, TriDiRA can enhance the quality of effective information in $r^*$ and $r \cap u$.


\section{Visualization}
\begin{figure}[t]
	\centering
	\includegraphics[width=90mm]{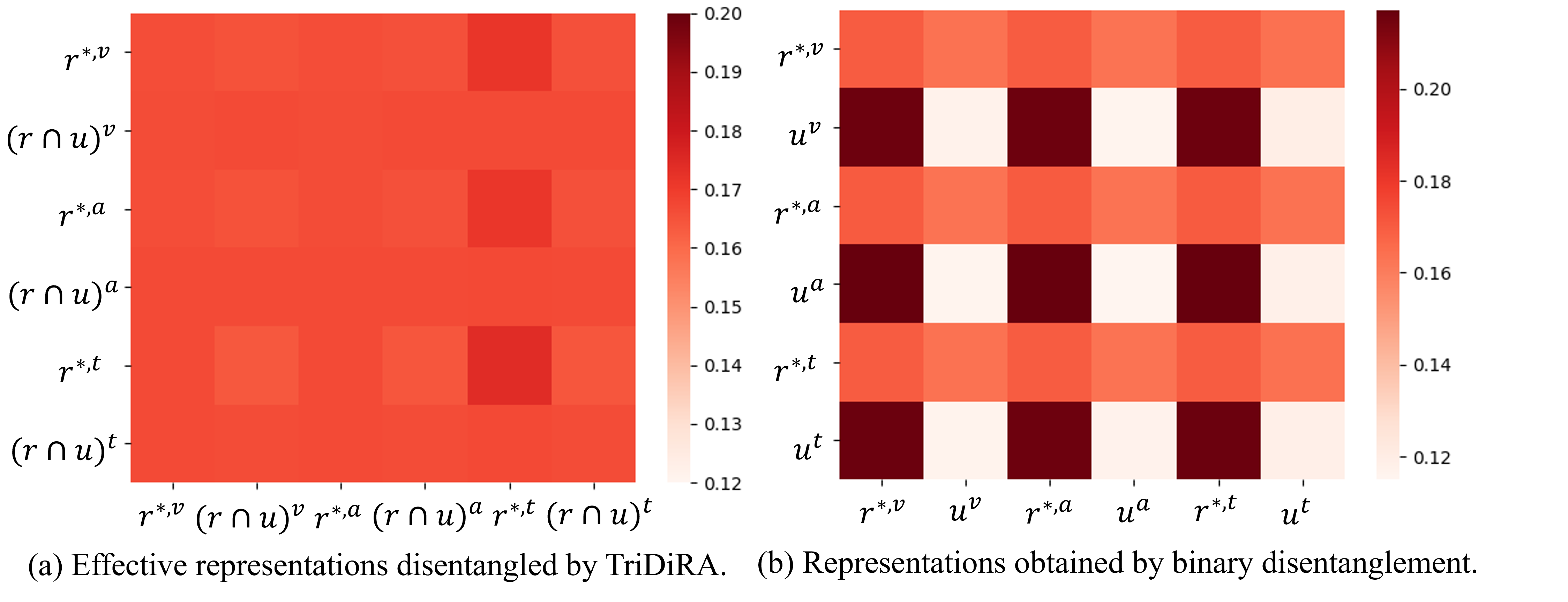}
	\caption{Average self-attention scores from the multi-head attention fusion module. The rows depict the queries, columns depict the keys. Essentially, each column represents the contribution of input representations to generate the output representations.}
	\label{fig:visual}
\end{figure}
In order to analyze the utility of the effective representations disentangled by TriDiRA, we visualized the attention process in the multi-head attention module. Here, we conducted two experiments, where the inputs to the multi-head attention were (a) the modality-invariant representations ($r^{*,m}$) and effective modality-specific representations ($(r \cap u)^m$) disentangled from each modality by TriDiRA, and (b) modality-invariant representations ($r^{*,m}$) and modality-specific representations ($u^m$) obtained by a binary disentanglement  model.

Multi-head attention module includes a self-attention procedure on the modality representations that enhances each input representation using a soft-attention combination of all representations.
Figure \ref{fig:visual} depicts the average attention distribution across the testing sets of (a) and (b) from CMU-MOSI. Each row represents a probability distribution of the corresponding representation. Each column illustrates the contribution of each representation to the final enhanced representation.

As can be observed in Fig. \ref{fig:visual}(a), each representation's contribution is relatively evenly distributed. In contrast, in Fig. \ref{fig:visual}(b), the modality-invariant representations have shown a significant influence, while the contributions among three $u^m$ are much less. This indicates that each representation in (a) contains effective information for the final prediction. However, in (b), ${u^m}$ provide less contributions to the final prediction. This further illustrates that ${u^m}$ of some samples may contain label-irrelevant information, which might downgrade the prediction. In this context, TriDiRA demonstrates superior performance compared to previous binary disentanglement methods. Additionally, the $r^{*,t}$ in (a) contributes to the prediction considerably higher than other components, possibly due to text providing more sentiment-related information.

\section{Complexity analysis}
\label{sec:complexity}

In the context of equivalent experimental conditions, we conducted a comparative analysis of the complexities of several methods on the large-scale dataset CMU-MOSEI. As shown in Table \ref{tab:complexity}, all the methods can converge within 15 epochs. One can see MISA exhibits a faster training speed per epoch, with TriDiRA following closely. However, as shown in Table 1 of the main paper, TriDiRA achieves the best performance. This underscores the success of TriDiRA in striking a balance between computational efficiency and task performance.

\begin{table}[t]
	\centering
	\caption{The complexity analysis.}
	\label{tab:complexity}
	\resizebox{62mm}{9.2mm}{
		\begin{tabular}[width=60mm]{l|cc}
			\hline
			Methods   & Training epoch & Time consumption per epoch  \\ \hline
			MISA  & 9 & 90$\pm$15s       \\
			MMIM & 12 & 245$\pm$15s     \\
			DMD &  15 & 410$\pm$15s   \\
			TriDiRA & 10 & 180$\pm$15s   \\ \hline
	\end{tabular}}
\end{table}

\section{Settings in TriDiRA}
\label{sec:setting}

\subsection{Datasets}
The detailed statistics of four datasets are listed in Table~\ref{tab:dataset}.

\begin{table}[]
	\centering
	\caption{The statistics of the  datasets.}
	\label{tab:dataset}
	\resizebox{75mm}{8.5mm}{
		\begin{tabular}[width=60mm]{l|cccc}
			\hline
			Dataset   & \quad Training Set\quad&\quad Valid Set \quad &\quad Test Set\quad &\quad All \quad   \\ \hline
			CMU-MOSI  & 1284      & 229       & 686      & 2199  \\
			CMU-MOSEI & 16326     & 1871      & 4659     & 22856 \\
			UR-FUNNY & 10598     & 2626      & 3290     & 16514 \\
			MELD     & 9989       & 1108      & 2610     & 13707 \\ \hline
	\end{tabular}}
\end{table}

\subsection{Experimental settings}
The experiments were conducted using the PyTorch framework, with Python version = 3.9 and PyTorch version = 2.0.

For each utterance, the feature dimensions of visual, audio and text are 64, 64 and 768, respectively. The hyperparameters are set to: $N^a = N^v = 2, N^t = N_S = 4, k=5$. For textual Transformers, the head number is set to 8. The head number of other Transformers is set to 4. All parameters are optimized by AdamW \cite{loshchilov2018fixing} with the weight decay of $5e-5$. We employed grid search to explore optimal combinations of weights. More detailed parameters are shown in Table \ref{tab:weights}. $d_{model}$ denotes the feature dimensions normalized by the 1-D convolution layer.

\begin{table}[]
	\centering
	\caption{Hyperparameter settings in TriDiRA.}
	\label{tab:weights}
	\begin{tabular}{lcccc}
		\hline
		& MOSI & MOSEI & UR-FUNNY & MELD\\
		\hline
		\multicolumn{5}{c}{Hyperparameters}  \\ \hline
		Learning rate & 8e-5  & 2e-5  & 1e-5 & 1e-5  \\
		Batch size & 64  & 24  & 32 & 32  \\
		$d_{model}$ & 128  & 256  & 256 & 256  \\ \hline
		\multicolumn{5}{c}{Weights of regularizations}  \\ \hline
		$\mathcal{L}_{sim}$ & 0.1  & 0.05  & 0.1 & 0.1  \\
		$\mathcal{L}_{ucorr}$& 0.8 & 0.5  & 0.3  & 0.3 \\
		$\mathcal{L}_{recon}$ &  0.2 & 0.15  & 0.2 & 0.2  \\
		$\mathcal{L}_{modality}$ & 0.05 & 0.03  & 0.05 & 0.05 \\
		$\mathcal{L}_h$  & 1.0 & 0.8 & 0.8 & 0.8\\
		$\mathcal{L}_{task}$ & 1.0 & 1.0 &1.0  & 1.0 \\
		\hline
	\end{tabular}
\end{table}

\subsection{The detailed structure of the feature extraction module}

\begin{figure*}[t]
	\centering
	\includegraphics[width=150mm]{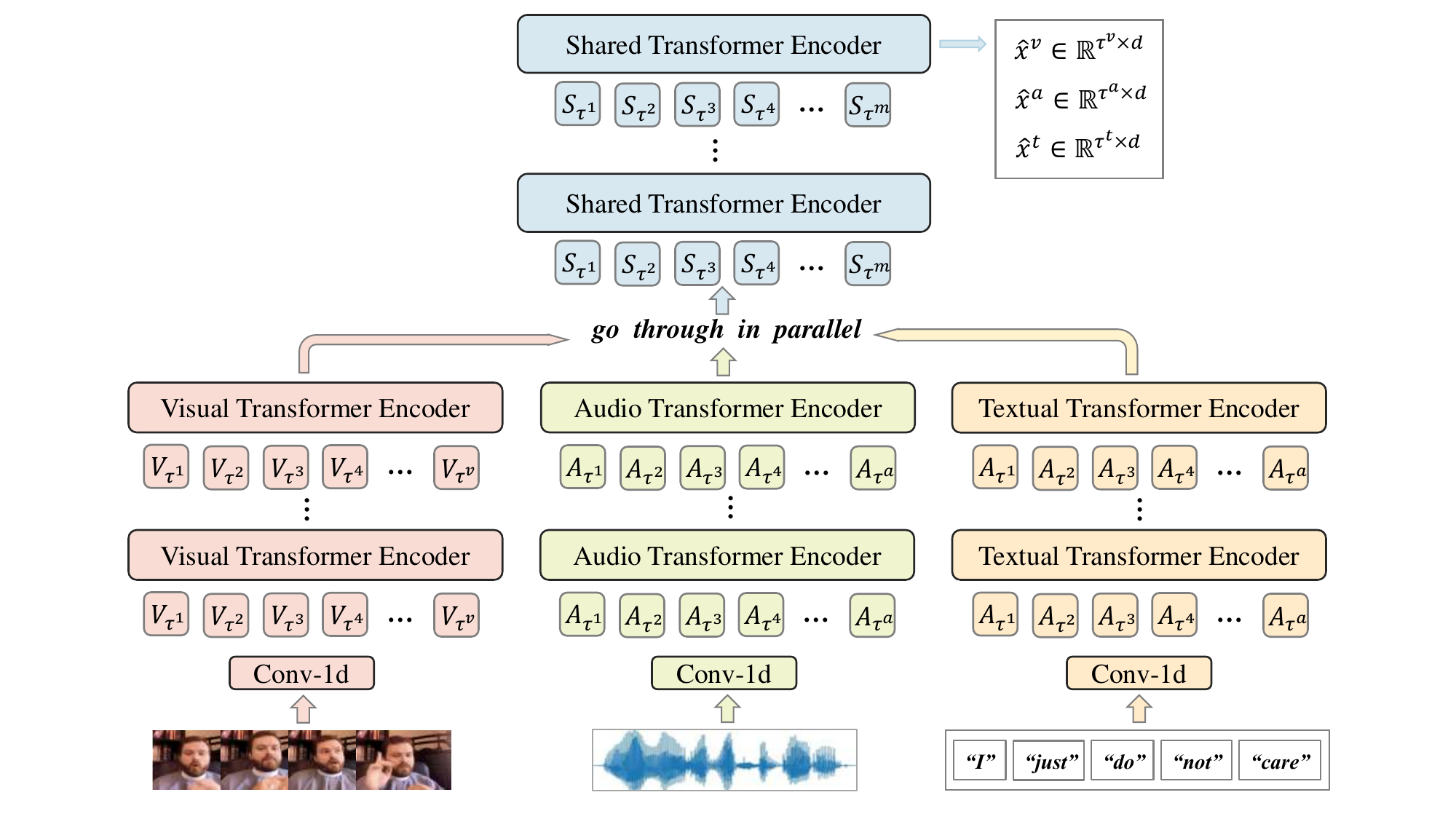}
	\caption{The detailed structure of the feature extraction module.}
	\label{fig:structure}
\end{figure*}

Initially, the feature $x^m$ is normalized by its corresponding 1-D convolution layers, and then fed into the  ${N^m}$-layer modality-specific Transformer to obtain ${{x}^{m'}}$. Afterwards, $\{{x}^{m'}\}$ are projected to the refined embeddings $\{\hat{x}^{m}\}$ in the same feature space by the shared modality-agnostic Transformer with ${N_S}$-layers. The detailed structure is shown in the Fig. \ref{fig:structure}.

\subsection{Training strategy}
\begin{figure*}[]
	\centering
	\includegraphics[width=130mm]{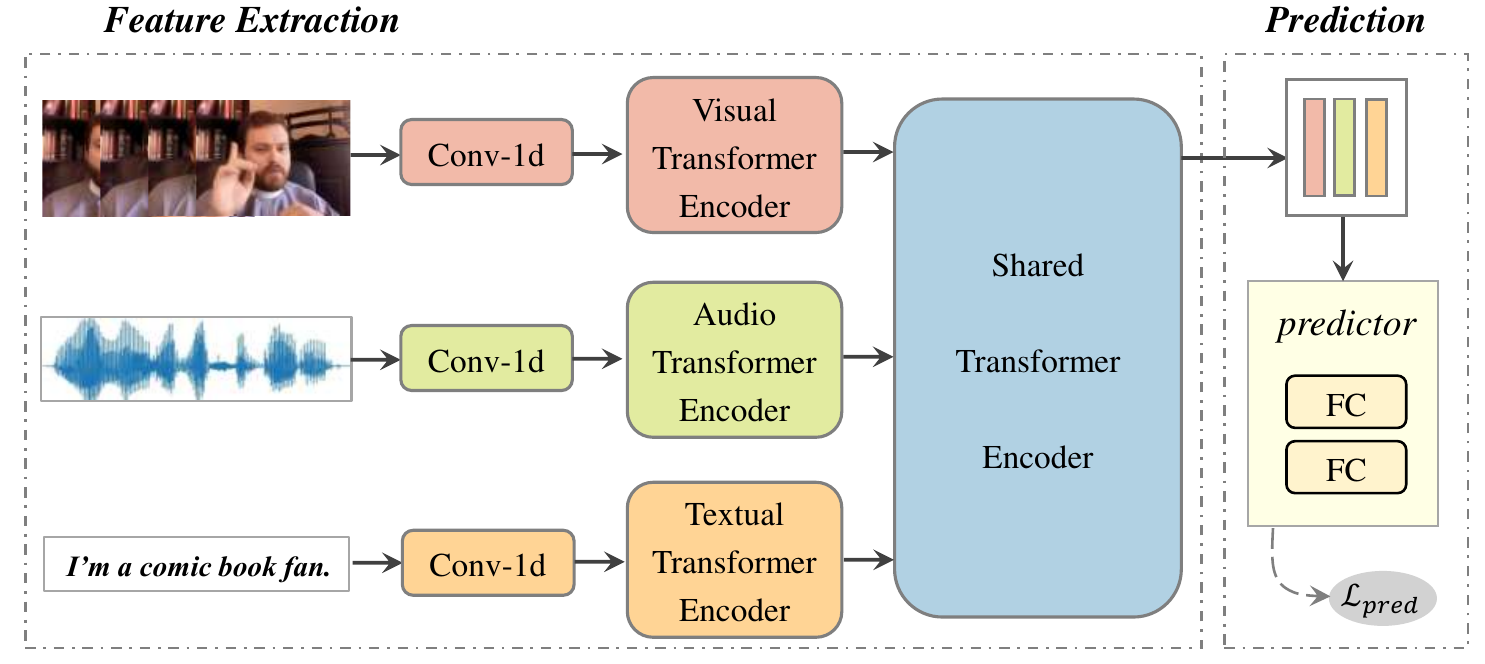}
	\caption{The first stage during the training.}
	\label{fig:warmup}
\end{figure*}

To address convergence issues during training on four datasets, a two-stage training strategy was adopted. In the first stage, the model without the DS module (Disentanglement module) was trained. In the second stage, the DS module and the multi-head attention module are plugged in for the following training. Additionally, concerning random seeds, the optimal weights of the first stage were applied while the average of five random seeds were employed in the second stage. The first stage during training is depicted in the Fig. \ref{fig:warmup}.

\section{The trends of regularizations}
\label{sec:loss}

\begin{figure*}[!h]
	\centering
	\includegraphics[width=120 mm]{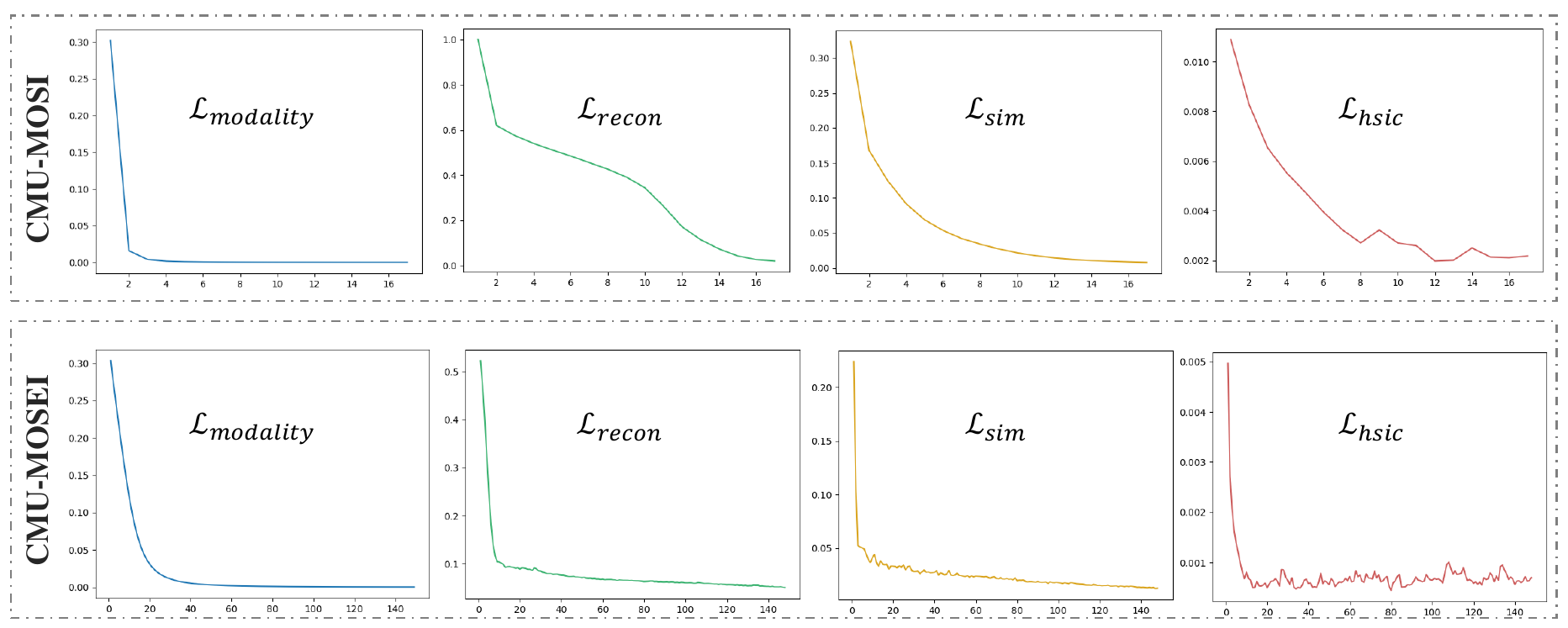}
	\caption{Trends in the regularization losses as training proceeds on CUM-MOSI. Similar trends are also observed on other datasets.}
	\label{fig:loss}
\end{figure*}

The losses act as measures to quantify how well the model has disentangled three  representations. We thus trace the losses as training proceeds in the training set of CMU-MOSI dataset. Figure \ref{fig:loss} shows the convergence curves of six different loss functions and the overall model loss. As can be observed, all losses demonstrate a decreasing trend with the number of epochs. And they can be converged within 40 epochs. This shows that the model is indeed learning the representations as per design.

\end{document}